\documentclass[letterpaper, 10 pt, conference]{ieeeconf}  

\IEEEoverridecommandlockouts                              

\usepackage{graphicx}
\usepackage{amsmath} 
\usepackage[english]{babel}
\usepackage[ruled]{algorithm2e}
\usepackage{floatrow}
\usepackage[caption=false]{subfig}
\usepackage{comment}
\usepackage{hyperref}
\usepackage{tabularx}
\newfloatcommand{capbtabbox}{table}[][\FBwidth]

\title{\LARGE \bf
Deep reinforcement learning oriented for real world dynamic scenarios
}

\author{Diego Martínez, Luis Riazuelo and Luis Montano$^{1}$
\thanks{$^{1}$The authors are with the Robotics, Perception and Real Time Group, Aragon Institute of Engineering Research (I3A), Universidad de Zaragoza, 50018 Zaragoza, Spain.
        {\tt\small diegomartinez,riazuelo,montano@unizar.es}}
}

\begin{document}

\maketitle
\thispagestyle{empty}
\pagestyle{empty}

\begin{abstract}
Autonomous navigation in dynamic environments is a complex but essential task for autonomous robots. Recent deep reinforcement learning approaches show promising results to solve the problem, but it is not solved yet, as they typically assume no robot kinodynamic restrictions, holonomic movement or perfect environment knowledge. Moreover, most algorithms fail in the real world due to the inability to generate real-world training data for the huge variability of possible scenarios. In this work, we present a novel planner, DQN-DOVS, that uses deep reinforcement learning on a descriptive robocentric velocity space model to navigate in highly dynamic environments. It is trained using a smart curriculum learning approach on a simulator that faithfully reproduces the real world, reducing the gap between the reality and simulation. We test the resulting algorithm in scenarios with different number of obstacles and compare it with many state-of-the-art approaches, obtaining a better performance. Finally, we try the algorithm in a ground robot, using the same setup as in the simulation experiments.

\end{abstract}

\section{Introduction}

Motion planning and navigation in dynamic scenarios is a complex problem that has not a defined solution. Traditional planners fail in environments where the map is mutable or obstacles are dynamic, leading to suboptimal trajectories or collisions. Those planners typically consider only the current obstacles' position measured by the sensors, without considering the future trajectories they may have.

New approaches that try to solve this issue include promising learning based methods. Nevertheless, they do not work properly in the real world: They do not consider robot kinodynamic constraints, only consider dynamic obstacles or assume perfect knowledge of the environment. Moreover, they would need huge real-world data to train the algorithms for the real world, and generating it is not possible.

We propose a planner that is able to navigate through dynamic and hybrid real-world environments. The planner is based on the Dynamic Object Velocity Space (DOVS) model, presented in \cite{lorente2018model}, which reflects the scenario dynamism information. In that work, the kinodynamics of the robot and the obstacles of the environment are used to establish the feasible velocities of the robot that do not lead to a collision. In our approach, the DOVS model is used in a new planner called DQN-DOVS, which utilizes deep reinforcement learning techniques. The planner applies the rich information provided by the DOVS as an input, taking advantage over other approaches that use raw sensor measurements and are not able to generalize. Once the agent learns how to interpret the DOVS, it is able to navigate in any scenario (it does not need real-world data of a huge variety of scenarios); and the training weights learned in the simulated world work as well in real-world environments without fine-tuning. In addition, it uses a dynamic window in the robot velocity space to define the set of actions available keeping robot kinodymanics.

\begin{figure}
    \centering
    \includegraphics[width=\textwidth]{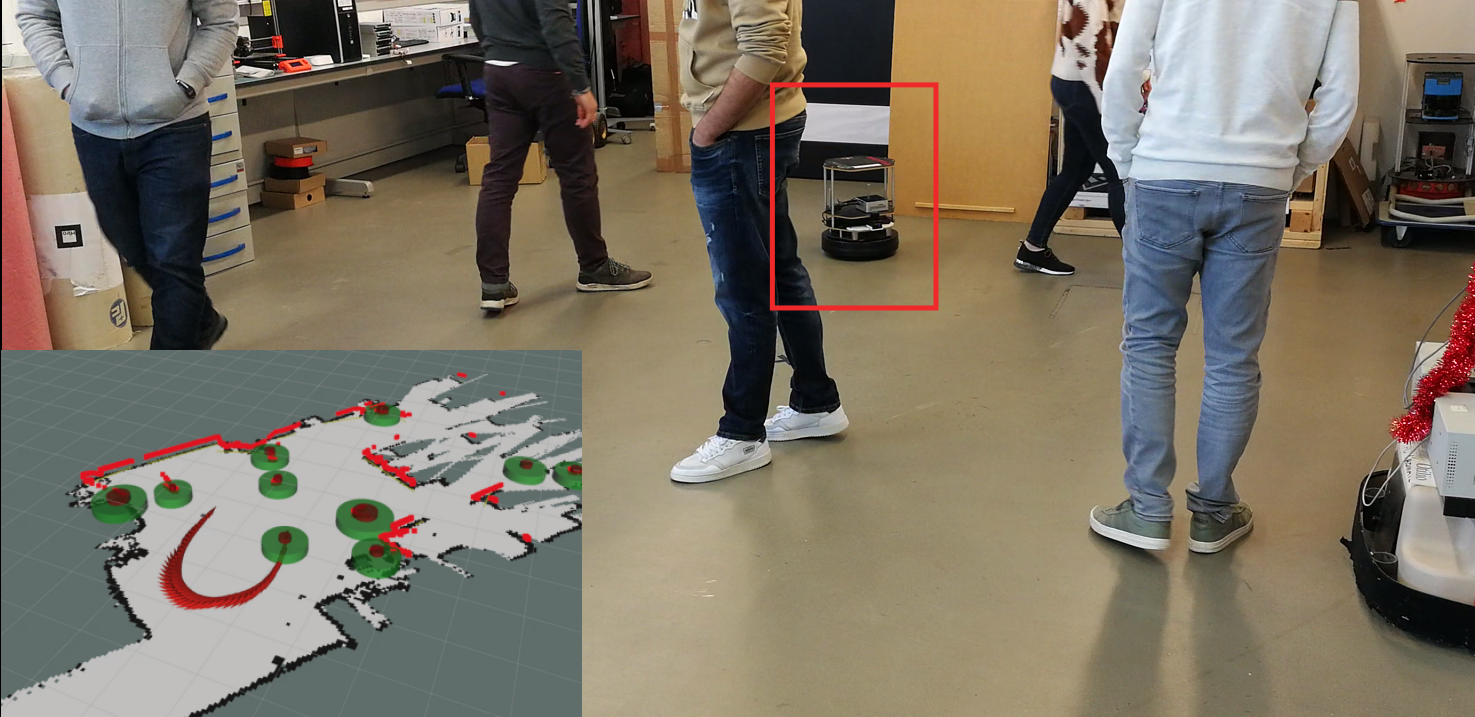}
    \caption{Scenario of a robot with static and dynamic obstacles and the RVIZ visualization of the scenario sensed.}
    \label{fig:main-fig}
\end{figure}

The DQN-DOVS algorithm is trained and tested in a real-world simulator, where all information is extracted from the sensor measurements, even the own robot localization. A comparison of the model with other planners of the state-of-the-art is also provided, as well as other experiments working with a real robot, like in Figure~\ref{fig:main-fig}. 

\section{Background}

\subsection{Related work}

Motion planners of static and continuous environments may not be used to deal with dynamic obstacles, as they lead to collisions and suboptimal trajectories. Some traditional approaches for dynamic environments include artificial potential fields \cite{qixin2006evolutionary}, are probability based \cite{van2012motion} or use a reciprocal collision avoidance (ORCA) \cite{van2008reciprocal}.

A big group of works are velocity space based. The \textit{Velocity Obstacle (VO)}, introduced in  \cite{fiorini1998motion},  refers to the set of velocities of the robot that could lead to collide with an obstacle that has a certain velocity in the near-future, which should not be chosen. Based on the VO concept,  the \textit{Dynamic Object Velocity Space (DOVS)} is defined in \cite{lorente2018model} as the velocity-time space for non-holonomic robots, which includes the unsafe robot velocities for all the obstacles and the time to collision for computing safe robot velocities in a time horizon. In the work, a planner based on strategies is also defined, the S-DOVS. A planner based on basic reinforcement learning on top of the DOVS is also proposed in \cite{mackay2022rl}, making decisions based on Q-values stored in tables.

Reinforcement learning is a method used to learn to estimate the optimal policy that optimizes the cumulative reward obtained in an episode. In \cite{mnih2015human}, the Q-values are estimated with a deep neural network, defining the first Deep Q-Network (DQN). Many extensions have been proposed to this original algorithm. Some works have proven the best performances of the state-of-the-art, including DQN with multiple modifications \cite{hessel2018rainbow}, distributed reinforcement learning \cite{horgan2018distributed} or actor-critic methods \cite{haarnoja2018soft}. The study presented in \cite{narvekar2020curriculum} shows that combining reinforcement learning with curriculum learning, could give useful results, specially to learn problems that could be too difficult to learn from scratch. 

Some works offer analysis of the importance of reinforcement learning in robot motion planning and the limitations of traditional planners in dynamic environments. Defining strategies for every situation that may be found in the real world is intractable, and reinforcement learning may be used to solve the decision-making problem, which is complex and has many degrees of freedom. \cite{kastner2021arena} proposes a deep reinforcement learning model that takes as the input of the model LIDAR measurements and the position of the goal, obtaining better results than conventional planners. 

The work described in \cite{chen2019crowd} (SARL) simulates a crowd and try to make the robot anticipate the crowd interactions with other robots and with each others, comparing its method with ORCA \cite{van2008reciprocal} and other two deep reinforcement learning methods: CADRL \cite{chen2017decentralized} and LSTM-RL \cite{everett2018motion}. ORCA fails in this crowded environment due to the need of the reciprocal assumption, and CADRL fails because it does not take into account the whole crowd, just a single pair for interaction. In the environment presented, both SARL and LSTM-RL have the best performance. In all of these approaches, the simulators used are non-realistic and no kinodynamic restrictions are considered. 

An example of an approach that considers the restrictions is \cite{patel2021dwa}, which combines deep reinforcement learning with DWA \cite{fox1997dynamic}, but only achieving a success rate of 0.54 in sparse dynamic scenarios.

\subsection{Dynamic Object Velocity Space (DOVS)}\label{sec:dovs}

The DOVS model presented in \cite{lorente2018model}, which models the dynamism of the environment, is used as a basis of this work. To build the model, the robot size is reduced to a point and the obstacles are enlarged with the robot radius (the final collision areas are the same). The area swept by each moving obstacle (collision band) is computed using the trajectory of the obstacles, which is assumed to be known or estimated from the sensor information. Then, the maximum and minimum velocities that can avoid a collision with that obstacle are calculated, repeating the process for every obstacle. Using that information with maximum and minimum velocities of the robot the limits of the \textit{Dynamic Object Velocity} (DOV) are obtained. The key of this model is that the set of free velocities is recomputed in every time step, so the obstacles' trajectory estimation needs to be precise only for the next few time steps.

\begin{figure}[h]
    \centering
    \includegraphics[width=0.9\textwidth]{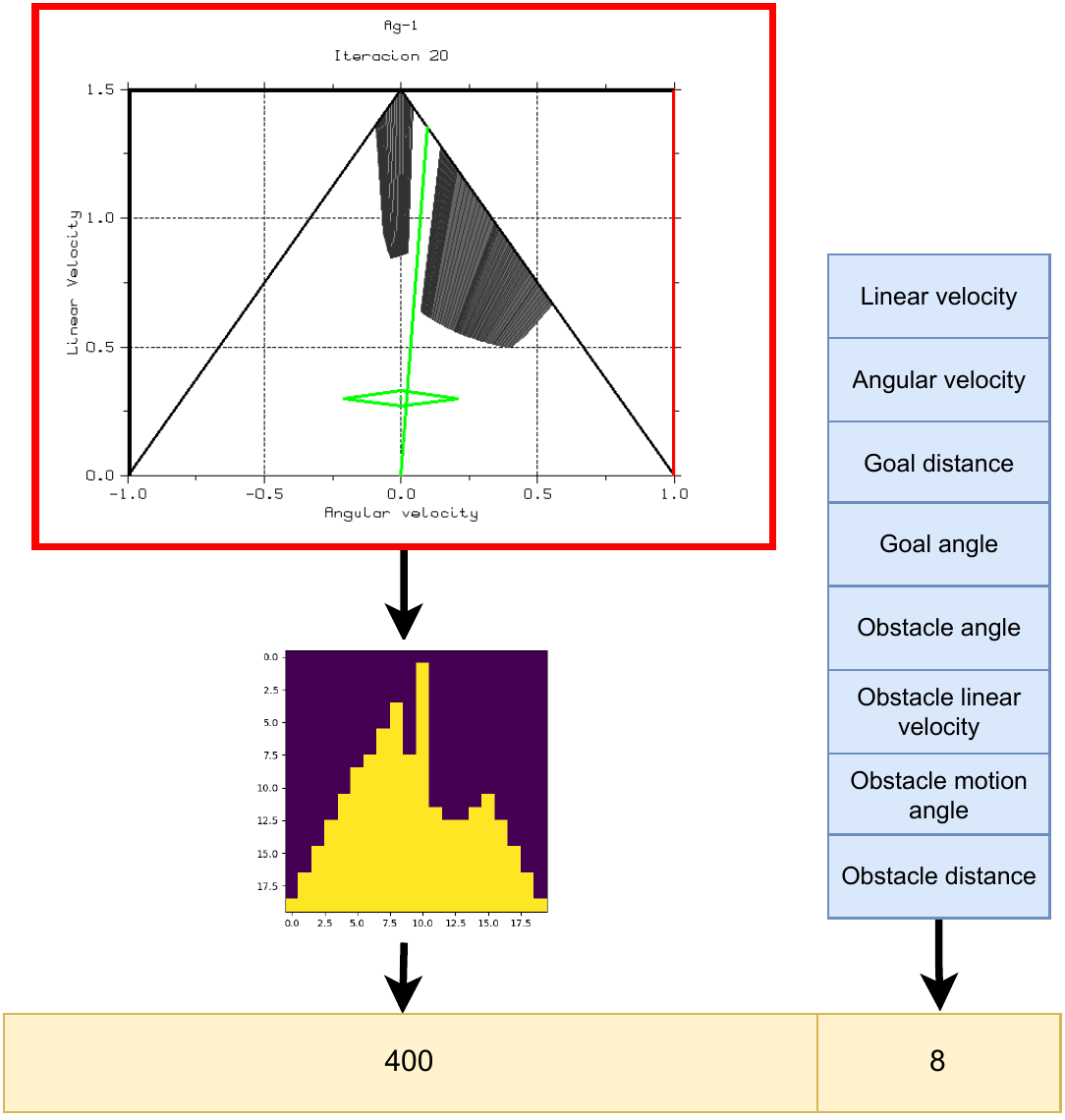}
    \caption{Representation of the state of the agent. The DOVS is in the red square and the extracted velocity grid below, concatenated to other 8 robot variables to construct the 408 elements input vector of the learning system.}
    \label{fig:DQN-state}
\end{figure}

 The velocities in the DOV are unsafe, as they lead to collisions in a time horizon, while the rest of the velocities are available to navigate. Velocities inside the DOV could be chosen if the following commands lead to a free velocity. The navigation in that space is achieved using a dynamic window that considers robot kinodynamics. The DOVS is built including all the information in the robot velocity space, and it may be represented as seen in Figure~\ref{fig:DQN-state}, inside a red square. Linear velocities $v$ are in the Y-axis, angular velocities $\omega$ in the X-axis, DOV are in black, the green rhombus is the dynamic window centered around the robot current velocity, the green line represents the velocities $(\omega,v)$ that lead to the goal following a circular trajectory ($radius=v/\omega$), and the big black triangle the differential-drive kinematic restriction (the robot may not go at maximum linear and angular velocity at the same time). In this way, all the information about the dynamism of the environment and the own robot needed for the robot motion planning is modeled.

\subsection{Contribution}

The works presented in the state of the art have some limitations when they are to be applied in real environments. Some of them use only the raw sensor measurements as the input or use some processed information, like obstacle position and velocities. The main problem with those approaches is the impossibility to generate appropriate real-world training data. They are trained in non-realistic simulators, and in different real world scenarios they would not do what to do. Furthermore, only few approaches do not use holonomic robots, and even less consider robot kinodynamic restrictions.

The contribution of this work is a deep reinforcement learning motion planner that:
\begin{itemize}
    \item Uses a very complete and descriptive information of the environment as the input, as it is the DOVS. The information from the obstacles is extracted with an obstacle tracker in the same simulation and the real world from the laser sensor measurements, and used to build the DOVS (with safe and unsafe velocities). Once the agent learns how to interpret the DOVS, it will be able to navigate in any kind of real-world scenario with the same trained weights (overcoming the unsolvable the problem of generating real-world data).
    \item Takes into account dynamic and static obstacles (people, robots, walls...).
    \item Is trained in a realistic simulator, considering occlusions and obstacle velocities estimation errors.
    \item Publishes differential-drive motion commands that take into account kinodynamic restrictions of the robot.
    \item Is able to brake considering the deceleration constraint when it is reaching the goal (others do not, as they they do not consider kinodynamic constraints).
    \item Receives all the information from sensors.
\end{itemize}

\section{Approach}

\subsection{State and action spaces}

The state should describe the environment. In our approach, the DOVS is used as the main part of the state to model the needed information of the obstacles in a fixed way. Using raw velocity information of obstacles would require training with obstacles with every possible shape, velocity or radius a robot could face, which is impossible; so using the DOVS is a big advantage. The information of safe and unsafe velocities are extracted from the DOVS as a 20x20 grid, assigning value of 1 to free velocities and -1 to obstacle velocities (DOV), to keep the relationships between the velocities in the velocity space. Other few variables are also added to the state, to describe the current robot situation and the information of the closest obstacle in case there is an imminent collision. The whole state is represented in Figure~\ref{fig:DQN-state}.

\begin{figure}[h]
    \centering
    \includegraphics[width=0.98\textwidth]{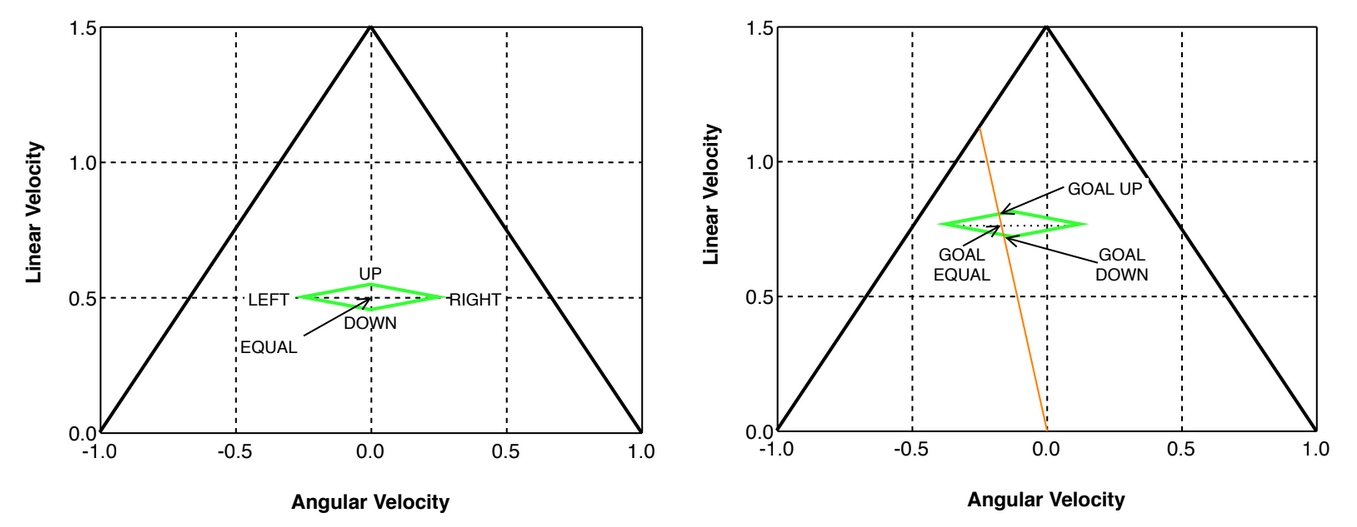}
    \caption{Action set in the DQN-DOVS algorithm. }
    \label{fig:action-set}
\end{figure}

The action space chosen is a discrete action set, relative to the current robot velocity and using the rhombus dynamic window (DW), to respect real robot kinodynamics. There are up to 8 available actions defined. 5 may be always chosen: The 4 corners of the DW (using maximum accelerations) and keeping the current velocity. The other three are only available if the velocities that lead to the goal are in DW: The two intersections of the line that leads to the goal with the dynamic window (maximum and minimum velocities for the next control period) and heading the goal with the current linear velocity. The whole set is represented in Figure~\ref{fig:action-set}. It is chosen as they are the actions that are more likely to be taken by natural trajectories, suitable for any differential drive robot. It is a discrete space instead of continuous to simplify the problem to improve training.

\subsection{DQN-DOVS}

The deep reinforcement learning algorithm chosen for the implementation is a Deep Q-Network with extensions, similar to the Rainbow DQN \cite{hessel2018rainbow}, as it offers results comparable to all the state-of-the-art in most of the problems with discrete action spaces. A decaying $\epsilon$-greedy strategy is used to balance exploration and exploitation.

The extensions used from the original DQN approach are Huber loss \cite{huber1964robust} (more robust to outliers), Double Q-learning \cite{van2016deep} (avoiding the overestimation of the Q-values), Prioritized Experience Replay, \cite{schaul2015prioritized} (sampling more often transitions the agent may learn more from), Dueling DQN \cite{wang2016dueling} (using the advantage function to compute the Q-values) and N-step bootstrapping \cite{sutton2018reinforcement} (bootstrapping the reward of several steps in the target value computation). Invalid actions (actions that lead to the goal when they are ineligible) are also taken into account in both the policy (what action to choose) and the target error computation to update the network weights. To do it, the invalid actions of each transition are stored in the replay buffer.

\begin{figure}[h]
    \centering
    \includegraphics[width=\textwidth]{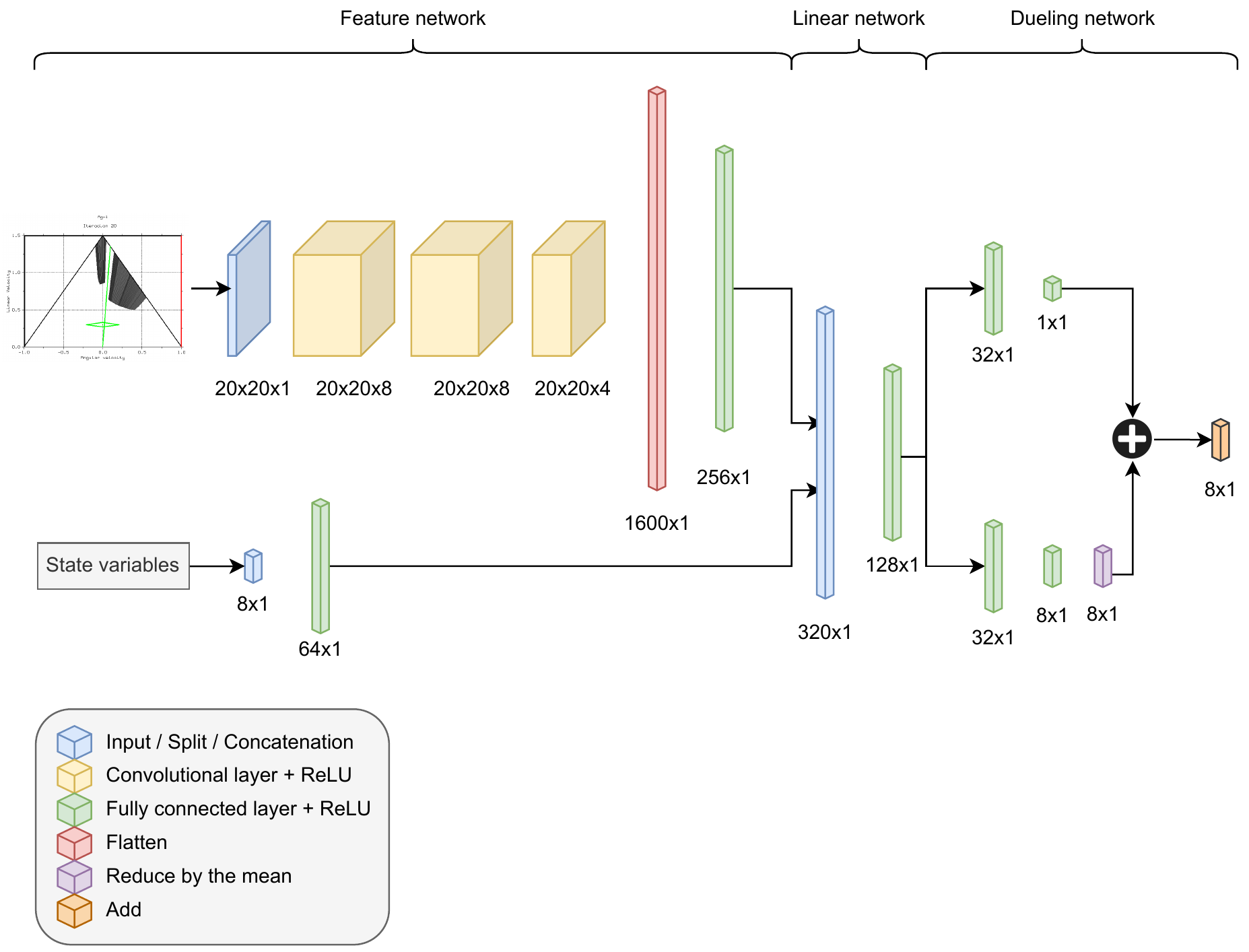}
    \caption{The structure of the network of the DQN-DOVS.}
    \label{fig:dqn-network}
\end{figure}

The Q-network is divided in three different parts sequentially connected: A feature network, a linear network and a dueling network. The feature network process separately the parts of the state. The DOVS image is fed into a convolutional network to extract the relationships among safe and unsafe velocities that are close to each other in the velocity space. The other 8 state variables are processed with a fully connected layer to exploit the relationships among them and increase their importance in the decision process. The outputs of both streams are concatenated into a linear network, which combines them. Finally, the dueling network computes the final Q-values of the actions by computing the state value and the advantage values in two streams. The structure of the network is shown in Figure~\ref{fig:dqn-network}.

\subsection{Reward function}

The goal of the agent is reaching and stopping in the goal while avoiding collisions in the shortest time possible. To achieve this behavior, the reward functions proposed in \cite{long2018towards} and \cite{sathyamoorthy2020densecavoid} are used as inspiration. It is defined with a simple equation that discriminates between terminal and non-terminal states:

\begin{equation}
   r^t =  \left\{
    \begin{array}{ll}
        r_{goal}, & d^t_{goal} < 0.15 \textnormal{ and }\\ & v^t < 0.2\\
        r_{collision}, & \textnormal{collision detected} \\
        -r_{dist}\Delta{d^t_{goal}} + r^t_{safedist}, & \textnormal{otherwise}
    \end{array}
    \right.
\end{equation}
The robot receives a reward of $r_{goal}$, set to 15, when it reaches and stops in the goal, using thresholds for the distance of the robot to the goal, $d^t_{goal}$, and for the linear velocity, $v^t$; and a negative reward of -15 when it collides ($r_{collision}$). Reward shaping is used to accelerate training in non-terminal states by encouraging the agent to get closer to the goal ($\Delta{d^t_{goal}}$ is the increment in the distance to the goal in consecutive time steps, $r_{dist}$ is set to 2.5), and by penalizing the agent if it is too close to an obstacle ($d^t_{obs}$ is the distance of the robot to the closest obstacle):

\begin{equation}
    r^t_{safedist} =  \left\{
    \begin{array}{ll}
        -0.1|0.2-d^t_{obs}|, & d^t_{obs} < 0.2\\
        0, & \textnormal{otherwise}
    \end{array}
    \right.
\end{equation}

\subsection{Training the network} \label{sec:curriculum}

The policy is trained in a modified version of the Stage simulator, which allows simulating robots similarly to the real world. An extended version of the work proposed by \cite{przybyla2017detection} is used to detect static and dynamic obstacles and estimate their position, radius, heading angle, angular and linear velocity in real time, from 2-D LIDAR measurements.Training in this kind of conditions makes the robot face occlusions or estimation errors already in training.

The training approach proposed in \cite{sathyamoorthy2020densecavoid} has been used as influence to design the curriculum learning strategy applied for training. In our work, the policy has been trained in a 8x8 m scenario delimited with walls; with random positions for the robot, its goal and the other obstacles (and random obstacle velocities). The stages used are shown in Table~\ref{tab:curriculum}, trying to make the agent progressively learn how to reach the goal, avoid static obstacles and avoid dynamic obstacles; to fine-tune everything in the last stage. The distance to the goal and number of obstacles is also progressively increased inside some stages; and the $\epsilon$ value is decayed or not, depending on whether the task to learn is new. 

\begin{table}[ht]
    \centering

    \begin{tabular}{|c|c|c|c|c|}
    \hline
        \textbf{Episodes} & \textbf{Distance} & \textbf{$\epsilon$} & \textbf{Obstacles} & \textbf{Type}\\
        \hline
        1000 & 1 to any & Decay & 0 & -- \\
        1000 & Any & Decay & 0 to 12 & Static \\        
        1000 & Any & 0.05 & 0 to 12 & Static \\        
        1000 & Any & Decay & 0 to 12 & Dynamic \\        
        1000 & Any & 0.05 & 0 to 12 & Dynamic \\        
        2500 & Any & 0.05 & Random & Both \\        
        \hline
    \end{tabular}
    \caption{Curriculum learning stages}

    \label{tab:curriculum}
\end{table}

\section{Results}

\subsection{Training details}\label{sec:training}

The agent was trained in the way presented in Section~\ref{sec:curriculum}. If the simulation took more than 500 time steps, the episode was finished (the robot gets stuck). The time step is set to 0.2 s. The dynamic obstacles had a predefined random linear and angular velocity each and may not avoid the agent. The network converges in about 10 hours in a computer with a Ryzen 7 5800x processor, a NVIDIA GeForce RTX 3060 graphics card and 64 GB of RAM. The key parameters used are a decaying learning rate of 0.0003 to 0.0001 with an Adam optimizer \cite{kingma2014adam}, discount factor of 0.97, n-step of 5 and a period of 100 to update the target network.

\subsection{Simulation evaluation}

To evaluate the model, different open source implementations of other methods have been used. We compare our DQN-DOVS method with S-DOVS \cite{lorente2018model}, the LSTM-RL (LSTM-RL-D) implementation of \cite{everett2018motion}, the SG-D3QN \cite{zhou2022robot}, and the implementations of ORCA, SARL, CADRL and LSTM-RL (LSTM-RL-H) offered in \cite{chen2019crowd} with the weights or the training scripts they provide. DQN-DOVS and S-DOVS consider every kind of kinodynamic restrictions, LSTM-RL-D and SG-D3QN only consider differential drive restrictions (but, for example, use infinite acceleration) and the rest do not consider any restriction and are holonomic. They are tested with the restrictions they consider.


\begin{figure}[h]
    \centering
    \subfloat[RVIZ and Simulator]{\includegraphics[width=0.45\textwidth]{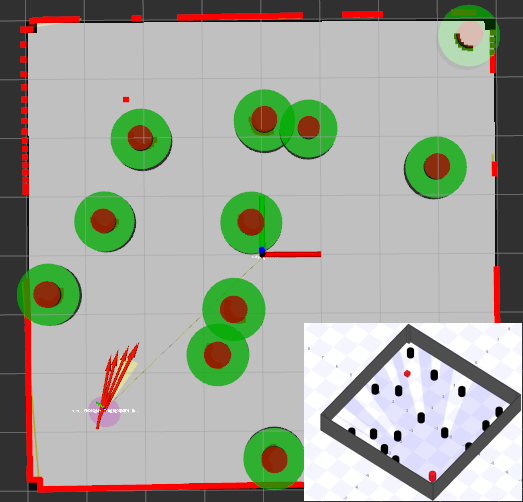}}
    \subfloat[Trajectories]{\includegraphics[width=0.45\textwidth]{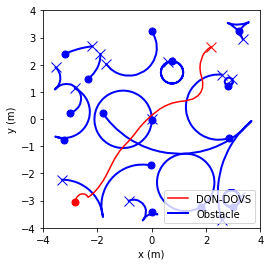}}
        \caption{Scenario of a DQN-DOVS red agent and 15 dynamic obstacles. The final position of the robot and obstacles are marked with a \textit{X}. Obstacle poses and velocities relative to the robot are computed from the simulated on board LIDAR sensor.}
        \label{fig:stage3d}
\end{figure}

The methods have been tested in 200 episodes of the same kind of scenarios and obstacles described in Section~\ref{sec:training}, varying the number of agents, from 1 to 15 with 85\% of the obstacles being dynamic (the same set of scenarios for every method). Although the scenarios may seem similar visually, what the robot perceives is completely different, depending on the obstacles density and velocities. The results obtained are shown in Table~\ref{tab:my_label}, with the success rate and the consumed time rate with respect to the DQN-DOVS.


\begin{table*}
\caption{Success rate of every method for different obstacles. In brackets, time rate with respect to the DQN-DOVS ($\frac{\textnormal{time of the method}}{\textnormal{time of DQN-DOVS}})$.}
\label{tab:my_label}
\centering
\begin{tabularx}{\textwidth}{|l |*9{>{\centering\arraybackslash}X|}@{}}
\hline
Obs & DQN- & S-DOVS & LSTM- & SG- & CADRL & ORCA & SARL & LSTM- \\
 & DOVS & & RL-D & D3QN & & & & RL-H\\
\hline
1 & \textbf{0.94} & 0.94(1.13) & \textbf{0.94}(0.85) & 0.90(0.44) & 0.93 & 0.93 & 0.94 & 0.94\\
2 & \textbf{0.96} & 0.91(1.13) & 0.92(0.72) & 0.81(0.43) & 0.92 & 0.90 & 0.91 & 0.92\\
3 & \textbf{0.89} & 0.85(1.04) & 0.84(0.84) & 0.77(0.41) & 0.82 & 0.87 & 0.84 & 0.83\\
4 & 0.80 & \textbf{0.81}(1.03) & 0.78(0.86) & 0.70(0.48) & 0.79 & 0.76 & 0.79 & 0.78\\
5 & \textbf{0.82} & 0.73(1.07) & 0.76(0.91) & 0.55(0.46) & 0.78 & 0.76 & 0.76 & 0.75\\
6 & \textbf{0.80} & 0.71(1.06) & 0.63(0.84) & 0.57(0.47) & 0.68 & 0.65 & 0.65 & 0.65\\
7 & \textbf{0.74} & 0.71(1.14) & 0.63(0.92) & 0.52(0.43) & 0.72 & 0.70 & 0.70 & 0.73\\
8 & \textbf{0.70} & 0.55(1.03) & 0.57(1.06) & 0.44(0.41) & 0.54 & 0.60 & 0.63 & 0.60\\
9 & \textbf{0.72} & 0.60(1.17) & 0.53(1.18) & 0.30(0.51) & 0.57 & 0.57 & 0.52 & 0.58\\
10 & \textbf{0.66} & 0.58(1.13) & 0.53(1.00) & 0.30(0.44) & 0.52 & 0.55 & 0.56 & 0.52\\
11 & \textbf{0.70} & 0.48(1.06) & 0.54(1.04) & 0.29(0.48) & 0.51 & 0.49 & 0.55 & 0.54\\
12 & \textbf{0.69} & 0.61(1.02) & 0.48(1.05) & 0.27(0.42) & 0.42 & 0.49 & 0.46 & 0.47\\
13 & \textbf{0.55} & 0.53(1.04) & 0.34(1.19) & 0.23(0.54) & 0.48 & 0.50 & 0.46 & 0.40\\
14 & \textbf{0.58} & 0.52(1.18) & 0.39(1.12) & 0.21(0.38) & 0.46 & 0.43 & 0.39 & 0.45\\
15 & 0.53 & \textbf{0.56}(1.00) & 0.27(1.37) & 0.27(0.42) & 0.45 & 0.45 & 0.41 & 0.47\\
\hline
\end{tabularx}
\end{table*}

The results show that the new DQN-DOVS algorithm outperforms the rest of the state-of-the-art algorithms; even though the DQN-DOVS agent may not go through the walls (some agents do), it takes into account acceleration constraints and needs to brake to stop in the goal. Moreover, it performs better even being differential drive (while LSTM-RL-H version performs better than LSTM-RL-D with the only difference of being holonomic).

One of the differences between the new DQN-DOVS and the other algorithms is they assume perfect information knowledge. The results show that our method is the one that deals the best with those limitations, by moving in a way that both reduces occlusions and estimation errors and is less risky. The results also show that the premise on which this work is based is correct. The models that use raw or barely processed information as the input of the reinforcement learning algorithm do not perform successfully when the environment changes, while using the pre-processed information as DOVS generalizes better to any scenario. The DQN-DOVS improves strategies based methods as S-DOVS by using deep reinforcement learning, and other learning methods by using the DOVS.

DQN-DOVS is faster than the other algorithm that consider kinodynamic restrictions (S-DOVS). Times of LSTM-RL-D are comparable, even though LSTM-RL-D assumes no acceleration constraints (and success rates are not close). The SG-D3QN algorithm perform very risky maneuvers and very short trajectories, because it assumes the obstacles are going to behave exactly like the ones seen in training (with no occlusions or estimation errors too), and that is why its success rates are so low and it is so fast, as well as its lack of acceleration constraints that makes it have maximum velocity at any time. The time with respect to the other methods is not comparable because they do not have constraints and are holonomic. An example of behavior of the DQN-DOVS agent in a random scenario may be seen in Figure~\ref{fig:stage3d}.

\subsection{Real robot experiments}

The whole system was integrated in a Turtlebot 2 platform with a NUC with Intel Core i5-6260U CPU and 8 GB of RAM. The sensor used is a 180º Hokuyo 2D-LIDAR. The approach taken from the beginning of the work was applying the simulation similarly as in the real world, using ROS. Thus, the same network weights and nodes used in simulation were used in the ground robot, adapting the obstacle tracker node to detect people and other objects and using AMCL for localization.

\begin{figure}[h]
    \centering
     \begin{tabular}[b]{c}
         \includegraphics[width=0.43\textwidth]{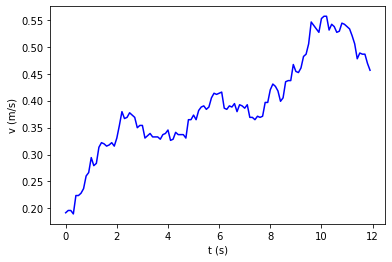}\\
         \footnotesize{(a) Linear velocity}
     \end{tabular}
     \hfill
     \begin{tabular}[b]{c}
         \includegraphics[width=0.43\textwidth]{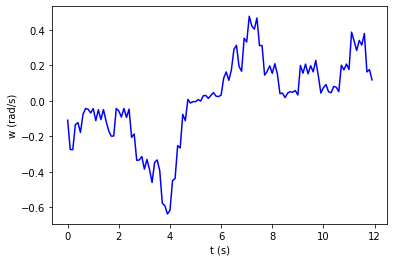} \\
         \footnotesize{(b) Angular velocity}
     \end{tabular}
          \begin{tabular}[b]{c}
         \includegraphics[width=0.32\textwidth]{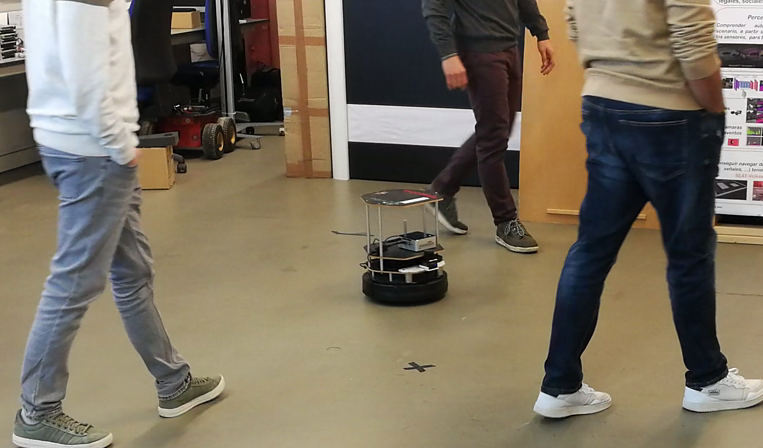}
         \includegraphics[width=0.32\textwidth]{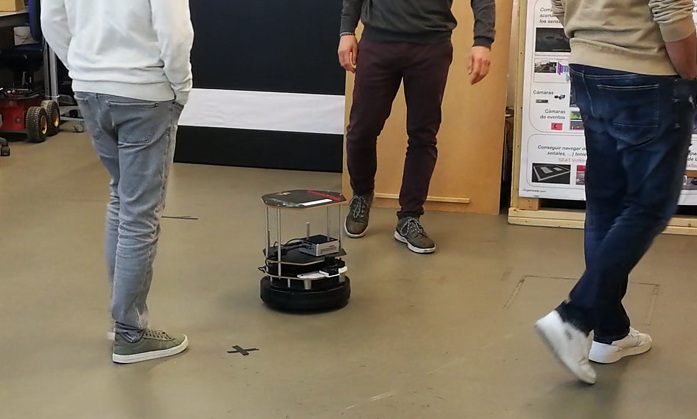}
         \includegraphics[trim={0.0cm 0.95cm 0.0cm 0.0cm},clip,width=0.32\textwidth]{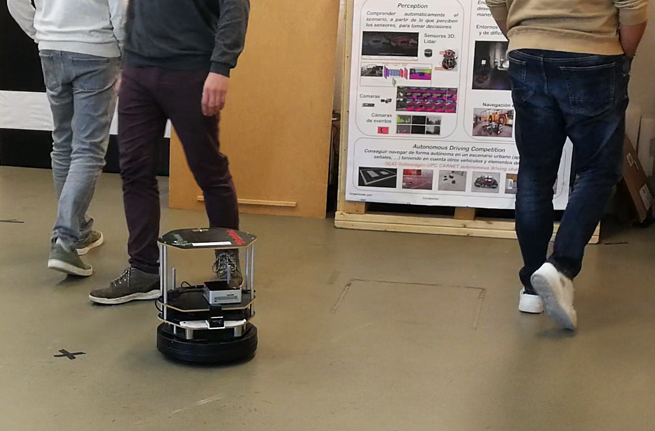} \\
         \footnotesize{(c) Pictures of the interval.}
     \end{tabular}
        \caption{Velocity measurements of the real robot in a short interval of time.}
        \label{fig:velocities}
\end{figure}

The experiments performed were setting the robot in a 8x8 m scenario where it had to navigate through different goals dynamically sent. Several people were wandering, acting as dynamic obstacles, and they were told not to look at the robot so that the avoidance was completely performed by it. In Figure~\ref{fig:velocities}, the robot velocity profiles during a time period in a experiment are shown, where the robot clearly respects the differential-drive kinodynamic restrictions, as it speeds up or decelerates respecting the acceleration constraints. In that specific situation, three moving obstacles surround the robot. The robot takes into account the obstacles trajectories and accelerates, turns to the right (second 3) and to the left (seconds 6, 11), and accelerates again, avoiding collisions successfully. The experiments performed show that the robot tries to predict obstacles trajectories in advance, and choose natural trajectories with maximum velocities to reach the goal, instead of avoiding obstacles clearly stopping and turning as other planners.

\section{Conclusion}

This work presents a motion planner and navigation algorithm for dynamic environments, that uses a differential drive robot, and the DOVS to model the environment as an input of a deep reinforcement learning technique to select the motion commands that lead the agent to reach the goal whilst avoiding collisions in the lower time possible. In addition, it proposes a training framework that uses curriculum learning in realistic scenarios to help the network to converge. The work is tested in random scenarios in a simulator where all the information is captured using a 2-D LIDAR sensor as in real life, outperforming existing methods that should have advantage, as they do not consider kinodynamic constraints of the robot and some of them use holonomic robots. The algorithm is also tested in a ground robot walking through pedestrians. Future work will include extending the model to collaborative collision avoidance with multi-robot navigation, trying more complex algorithms, and adapting the model for 3-D navigation and UAV (the model and the approach is extensible for 3-D spaces).

\section*{Acknowladgement}

This work was partially supported by the Spanish projects MCIN/AEI/PID2019-105390RB-I00, and~Aragon Government\_FSE-T45\_20R.

\bibliographystyle{IEEEtran}
\bibliography{IEEEabrv,main}

\end{document}